\documentclass[10pt,twocolumn,letterpaper]{article}
\usepackage{cvpr}
\usepackage{graphicx}
\usepackage{amsmath}
\usepackage{amssymb}
\usepackage{booktabs}
\usepackage{multirow}
\usepackage{dsfont}
\usepackage{pifont}
\usepackage{xcolor}
\usepackage{colortbl}
\usepackage{makecell}
\usepackage[accsupp]{axessibility}

\newcommand{\authorskip}{\hspace{8mm}}

\usepackage[pagebackref,breaklinks,colorlinks]{hyperref}

\usepackage[capitalize]{cleveref}
\crefname{section}{Sec.}{Secs.}
\Crefname{section}{Section}{Sections}
\Crefname{table}{Table}{Tables}
\crefname{table}{Tab.}{Tabs.}

\def\cvprPaperID{7927} 
\def\confName{CVPR}
\def\confYear{2023}

\begin{document}

\title{
DeepSolo: Let Transformer Decoder with Explicit Points Solo for Text Spotting
}

\author{
  Maoyuan Ye$^{1}$\thanks{Equal contribution. \dag Corresponding author. This work was done during Maoyuan Ye's internship at JD Explore Academy.} \authorskip Jing Zhang$^{2}$\footnotemark[1] \authorskip Shanshan Zhao$^{3}$ \authorskip Juhua Liu$^{1\dag}$\\
  Tongliang Liu$^{2}$ \authorskip Bo Du$^{1\dag}$ \authorskip Dacheng Tao$^{3,2}$\\
  $^{1}$Wuhan University \quad $^{2}$The University of Sydney \quad $^{3}$JD Explore Academy\\
  {\tt\small \{yemaoyuan, liujuhua, dubo\}@whu.edu.cn, \{jing.zhang1, tongliang.liu\}@sydney.edu.au,}\\
  {\tt\small \{sshan.zhao00, dacheng.tao\}@gmail.com}
}

\maketitle

\begin{abstract}
   End-to-end text spotting aims to integrate scene text detection and recognition into a unified framework. Dealing with the relationship between the two sub-tasks plays a pivotal role in designing effective spotters. Although Transformer-based methods eliminate the heuristic post-processing, they still suffer from the synergy issue between the sub-tasks and low training efficiency. In this paper, we present \textbf{DeepSolo}, a simple DETR-like baseline that lets a single \textbf{De}coder with \textbf{E}xplicit \textbf{P}oints \textbf{Solo} for text detection and recognition simultaneously. Technically, for each text instance, we represent the character sequence as ordered points and model them with learnable explicit point queries. After passing a single decoder, the point queries have encoded requisite text semantics and locations, thus can be further decoded to the center line, boundary, script, and confidence of text via very simple prediction heads in parallel. Besides, we also introduce a text-matching criterion to deliver more accurate supervisory signals, thus enabling more efficient training. Quantitative experiments on public benchmarks demonstrate that DeepSolo outperforms previous state-of-the-art methods and achieves better training efficiency. In addition, DeepSolo is also compatible with line annotations, which require much less annotation cost than polygons. The code is available at \href{https://github.com/ViTAE-Transformer/DeepSolo}{DeepSolo}.
\end{abstract}

\section{Introduction}
\label{sec:intro}
Detecting and recognizing text in natural scenes, \textit{a.k.a.} text spotting, has drawn increasing attention due to its wide range of applications \cite{ye2014text,zhang2020empowering,chen2021text,long2021scene}, such as autonomous driving \cite{zhang2021character} and intelligent navigation \cite{desouza2002vision}. 
How to deal with the relationship between detection and recognition is a long-standing problem in designing the text spotting pipeline and has a significant impact on structure design, spotting performance, training efficiency, and annotation cost, etc.

\begin{figure}[!t]
    \centering
    \includegraphics[width=\linewidth]{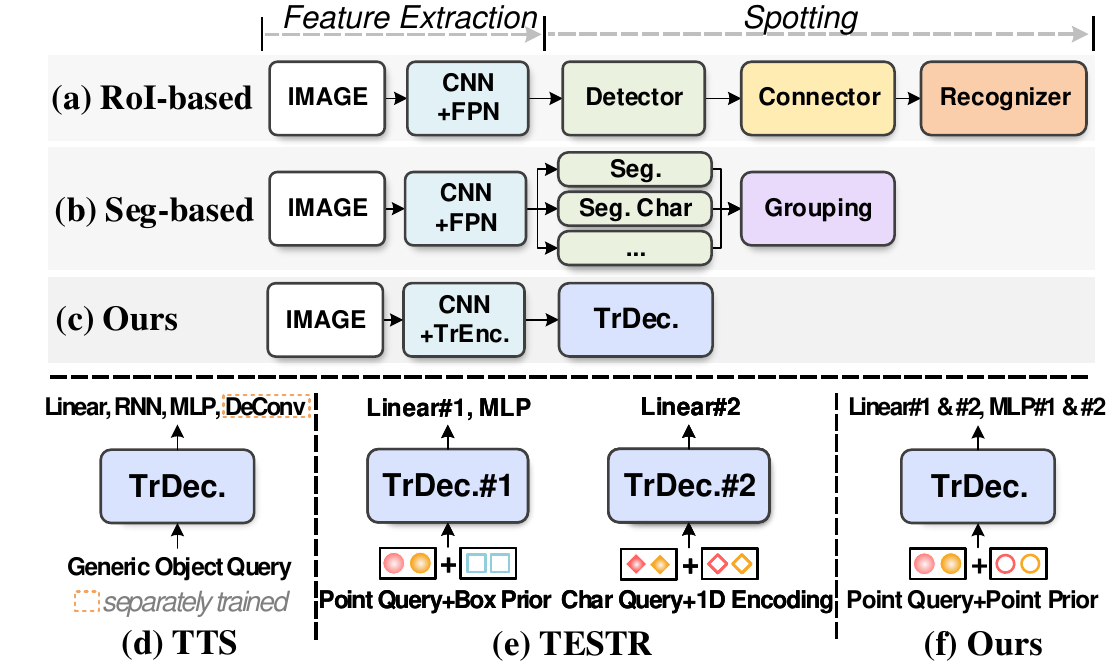}
    \caption{Comparison of pipelines and query designs. TrEnc. (TrDec.): Transformer encoder (decoder). Char: character.}
    \label{fig:1}
    \vspace{-6mm}
\end{figure}

Most pioneering end-to-end spotting methods \cite{he2018end,liu2018fots,liu2020abcnet, liu2021abcnet,wang2021pan++,lyu2018mask,liao2020mask,ronen2022glass,qin2019towards} follow a detect-then-recognize pipeline, which first detects text instances and then exploits Region-of-Interest (RoI) based connectors to extract features within the detected area, finally feeds them into the following recognizer (\cref{fig:1}a). Although these methods have achieved great progress, there are two main limitations. 1) An extra connector for feature alignment is indispensable. Moreover, some connectors require polygon annotations, which are not applicable when only weak annotations are available. 2) Additional efforts are desired to address the synergy issue \cite{zhong2021arts, huang2022swintextspotter} between the detection and recognition modules. In contrast, the segmentation-based methods \cite{xing2019convolutional,wang2021pgnet} try to isolate the two sub-tasks and complete spotting in a parallel multi-task framework with a shared backbone (\cref{fig:1}b).
Nevertheless, they are sensitive to noise and require grouping post-processing to gather unstructured components.

Recently, Transformer \cite{vaswani2017attention} has improved the performance remarkably for various computer vision tasks \cite{dosovitskiy2020image,liu2021swin,liu2022swin,wang2021pyramid,touvron2021training,xu2021vitae,zhang2022vitaev2,du2022i3cl,peng2021conformer,li2022panoptic}, including text spotting~\cite{huang2022swintextspotter,zhang2022text,peng2022spts,kittenplon2022towards}. Although the spotters \cite{zhang2022text,kittenplon2022towards} based on DETR \cite{carion2020end} can get rid of the connectors and heuristic post-processing, they lack efficient joint representation to deal with scene text detection and recognition, \eg, requiring an extra RNN module in TTS \cite{kittenplon2022towards} (\cref{fig:1}d) or exploiting individual Transformer decoder for each sub-task in TESTR \cite{zhang2022text} (\cref{fig:1}e). The generic object query exploited in TTS fails to consider the unique characteristics of scene text, \eg, location and shape. While TESTR uses point queries with box positional prior that is coarse for point predicting, and the queries are different for detection and recognition, introducing unexpected heterogeneity. Consequently, these designs have a side effect on the performance and training efficiency \cite{ye2022dptext}.

In this paper, we propose a novel query form based on explicit point representations of text lines. Built upon it, we present a succinct DETR-like baseline that lets a single \textbf{De}coder with \textbf{E}xplicit \textbf{P}oints \textbf{Solo} (dubbed \textbf{DeepSolo}) for detection and recognition simultaneously (\cref{fig:1}c and \cref{fig:1}f). Technically, for each instance, we first represent the character sequence as ordered points, where each point has explicit attributes of position, offsets to the top and bottom boundary, and category. Specifically, we devise top-$K$ Bezier center curves to fit scene text instances with arbitrary shape and sample a fixed number of on-curve points covering characters in each text instance. Then, we leverage the sampled points to generate positional queries and guide the learnable content queries with explicit positional prior. Next, we feed the image features from the Transformer encoder and the point queries into a single Transformer decoder, where the output queries are expected to have encoded requisite text semantics and locations. Finally, we adopt several very simple prediction heads (a linear layer or MLP) in parallel to decode the queries into the center line, boundary, script, and confidence of text, thereby solving detection and recognition simultaneously. Besides, we introduce a text-matching criterion to leverage scripts and deliver more accurate supervisory signals, thus further improving training efficiency.

In summary, the main contributions are three-fold: \textbf{1)} We propose DeepSolo, \ie, a succinct DETR-like baseline with a single Transformer decoder and several simple prediction heads, to solve text spotting efficiently. \textbf{2)} We propose a novel query form based on explicit points sampled from the Bezier center curve representation of text instance lines, which can efficiently encode the position, shape, and semantics of text, thus helping simplify the text spotting pipeline. \textbf{3)} Experimental results on public datasets demonstrate that DeepSolo is superior to previous representative methods in terms of spotting accuracy, training efficiency, and annotation flexibility.

\section{Related Works}
\subsection{Connector-based Text Spotters}
Recent works focus on developing end-to-end spotters. Most of them craft RoI \cite{he2017mask} based or Thin-Plate Splines (TPS) \cite{bookstein1989principal} based connectors to bridge detection and recognition modules. In \cite{he2018end, liu2018fots,feng2019textdragon}, given the detection results, RoI-Align or its variant is exploited to extract text features for the following recognizer. Mask TextSpotter series \cite{lyu2018mask, liao2020mask, liao2021mask} conduct character segmentation in recognition. GLASS \cite{ronen2022glass} devises a plug-in global-to-local attention module with Rotated-RoIAlign for strong recognition. In comparison, \cite{qiao2020text,wang2020all} use TPS to rectify curve texts with control points. To better rectify curve texts, ABCNet series \cite{liu2020abcnet,liu2021abcnet} propose the BezierAlign module using parameterized Bezier curve. Although these methods successfully bridge text detection and recognition, they ignore the synergy issue \cite{zhong2021arts, huang2022swintextspotter} between the two tasks. To overcome this dilemma, SwinTextSpotter \cite{huang2022swintextspotter} proposes a tailored Recognition Conversion module to back-propagate recognition information to the detector. Although the above methods have achieved remarkable progress, they require an extra RoI-based or TPS-based connector. Since some connectors require polygon annotations, the methods may be not applicable to scenarios with only weak position annotations (\eg, lines). Moreover, a more effective and simpler solution is also expected to address the synergy issue.

\begin{figure*}[!t]
    \centering
    \includegraphics[width=1\linewidth]{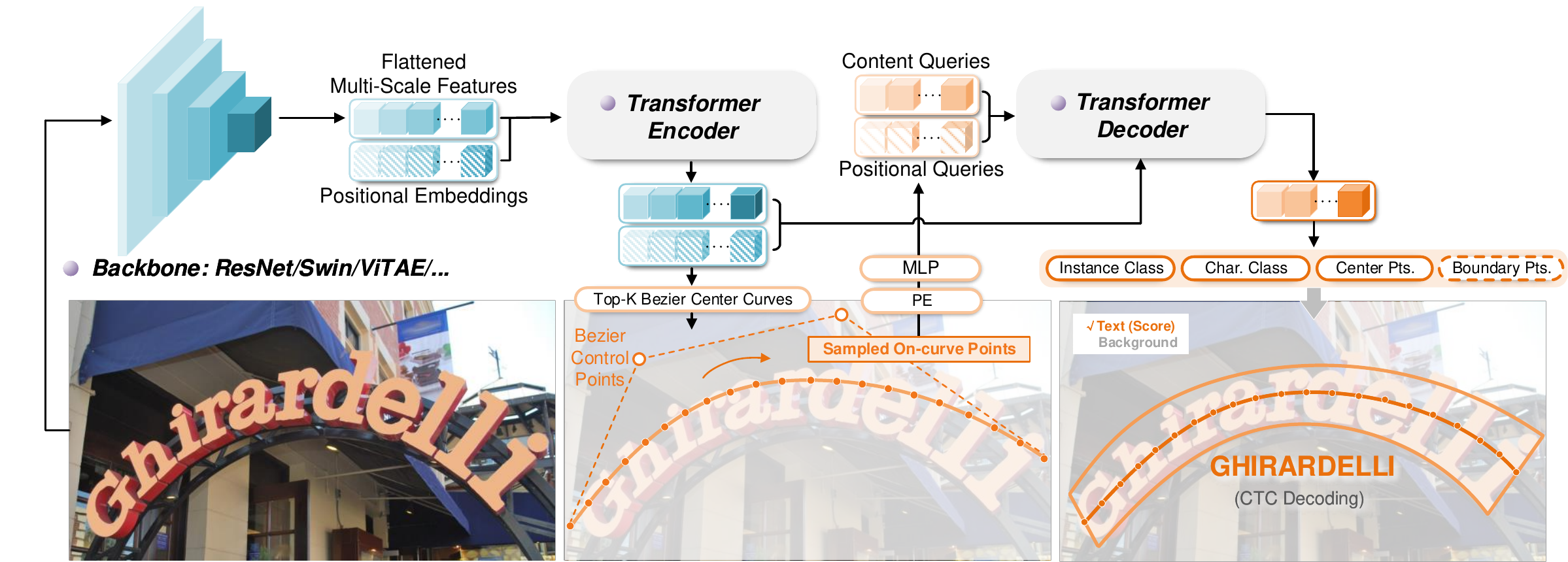}
    \caption{The overall architecture of DeepSolo. We propose an explicit query form based on the points sampled from the Bezier center curve representation of text, solving spotting with a single decoder and simple prediction heads in a unified and concise framework.}
    \label{fig:model}
\end{figure*}

\subsection{Connector-free Text Spotters}
\label{subsubsec:connector-free}
To get rid of the extra connector, segmentation-based methods \cite{xing2019convolutional,wang2021pgnet} adopt parallel branches to segment characters and instances. However, they tend to be vulnerable to noise and rely on heuristic grouping. In contrast, SRSTS \cite{wu2022decoupling} estimates positive anchors for each instance, then adopts a text shape segmentation branch and a self-reliant sampling recognition branch in parallel to effectively decouple detection and recognition. Nevertheless, it needs NMS for post-processing. 

Inspired by DETR family \cite{carion2020end, zhu2020deformable}, recent works \cite{kittenplon2022towards,peng2022spts,zhang2022text} explore the Transformer framework without RoI-Align and complicated post-processing. TESTR \cite{zhang2022text} adopts two parallel Transformer decoders for detection and recognition. TTS \cite{kittenplon2022towards} adds an RNN recognizer into Deformable-DETR \cite{zhu2020deformable} and shows its potential on weak annotations. SPTS \cite{peng2022spts} follows the sequence prediction paradigm \cite{chen2021pix2seq} and demonstrates that using single-point or even script-only annotations is feasible and promising. 

Although these methods unlock the potential of Transformer in text spotting, there are still some limitations. For example, the vanilla or individual queries used in TTS and TESTR cannot efficiently encode text features (\eg, location, shape, and semantics), affecting the training efficiency \cite{ye2022dptext} and even increasing the complexity of the pipeline. The box annotations used in TTS might not be ideal \cite{peng2022spts} for scene text since the box contains a certain portion of background regions and even other texts, thus introducing extra noise. SPTS, TTS, and our work provide more comprehensive but different supervision solutions for the community.

\section{Methodology}
\label{sec:method}
In this paper, we propose \textbf{DeepSolo} for text spotting. DeepSolo aims at achieving detection and recognition simultaneously and efficiently with a single Transformer decoder by digging into the close relationship between them.

\subsection{Overview}
\label{sec:overview}
\noindent\textbf{Preliminary.} Bezier curve, firstly introduced into text spotting by ABCNet \cite{liu2020abcnet}, can flexibly fit the shape of scene text with a certain number of control points. Different from ABCNet which crops features with BezierAlign, we explore the distinctive utilization of Bezier curve in the DETR framework. More details of the Bezier curve generation can be referred to \cite{liu2020abcnet, liu2021abcnet}. Given four Bezier control points\footnote{Provided by ABCNet \cite{liu2020abcnet}} for each side (top and bottom) of a text instance, we simply compute Bezier control points for the center curve by averaging the corresponding control points on top and bottom sides. Then, a fixed number of $N$ points are uniformly sampled on the center, top, and bottom curves, serving as the ground truth. Note that the order of the points is in line with the text reading order.

\noindent\textbf{Model Architecture}. The overall architecture is illustrated in \cref{fig:model}. After receiving the image features from the encoder equipped with deformable attention \cite{zhu2020deformable}, Bezier center curve proposals and their scores are generated. Then, the top-$K$ scored ones are selected. For each selected curve proposal represented by four Bezier control points, $N$ points are uniformly sampled on the curve. The coordinates of these points are encoded as positional queries and added to the learnable content queries, forming the composite queries. Next, the composite queries are fed into the decoder to gather useful text features via deformable cross-attention. Following the decoder, four simple parallel heads are adopted, each of which is responsible for solving a specific task. 

\subsection{Top-$K$ Bezier Center Curve Proposals}
\label{sec:top-k proposal}
Different from the box proposal adopted in previous works \cite{zhu2020deformable, zhang2022text, jia2022detrs, ye2022dptext, zhang2022dino}, which has the drawbacks in representing text with arbitrary shape, we design a simple Bezier center curve proposal scheme from the text center line perspective. It can efficiently fit scene text and distinguish one from others, which also makes the usage of line annotations possible. Specifically, given the image features from the encoder, on each pixel of the feature maps, a 3-layer MLP (8-dim in the last layer) is used to predict offsets to four Bezier control points, determining a curve that represents one text instance. Let $i$ index a pixel from features at level $l \in \left\{1,2,3,4\right\}$ with 2D normalized coordinates $\hat{p}_i = \left(\hat{p}_{ix}, \hat{p}_{iy}\right) \in \left[0,1\right]^2$, its corresponding Bezier control points $BP_i = \{\bar{bp}_{i_0}, \bar{bp}_{i_1}, \bar{bp}_{i_2}, \bar{bp}_{i_3}\}$ are predicted:
\begin{equation}
    \bar{bp}_{i_j} = (\sigma(\Delta p_{ix_j} + \sigma^{-1}(\hat{p}_{ix})), \sigma(\Delta p_{iy_j} + \sigma^{-1}(\hat{p}_{iy}))), \label{eq_1}
\end{equation}
where $j \in \{0,1,2,3\}$, $\sigma$ is the sigmoid function. The MLP head only predicts the offsets $\Delta p_{i\{x_0,y_0,\ldots,x_3,y_3\}}$. Moreover, we use a linear layer for text or not text classification. And top-$K$ scored curves are selected as the proposals.

\subsection{Point Query Modeling for Text Spotting}
\noindent \textbf{Query Initialization.} Given the top-$K$ proposals $\hat{BP}_k (k \in \{0,1,\ldots,K-1\})$, we uniformly sample $N$ points on each curve according to Bernstein Polynomials \cite{lorentz2013bernstein}. Here, we get normalized point coordinates $Coords$ of shape $K \times N \times 2$ in each image. The point positional queries $\textbf{P}_{q}$ of shape $K \times N \times 256$ are generated by:
\begin{equation}
    \textbf{P}_{q} = MLP(PE(Coords)), \label{eq_2}
\end{equation}
where $PE$ represents the sinusoidal positional encoding function. Following \cite{liu2022dab}, we also adopt a 2-layer $MLP$ head with the $ReLU$ activation for further projection. On the other side, we initialize point content queries $\textbf{C}_{q}$ using learnable embeddings. Then, we add $\textbf{P}_{q}$ to $\textbf{C}_{q}$ to get the composite queries $\textbf{Q}_q$:
\begin{equation}
    \textbf{Q}_q = \textbf{C}_{q} + \textbf{P}_{q}. \label{eq_3}
\end{equation}
We empirically exploit unshared point embeddings for $\textbf{C}_{q}$, \ie, $N$ point content queries of shape $N \times 256$ in one text instance are used for each of the $K$ proposals.

\noindent \textbf{Query Update in the Decoder.} After obtaining the composite queries $\textbf{Q}_q$, we feed them into the Transformer decoder. We follow previous works \cite{dong2021visual,zhang2022text,ye2022dptext} to firstly mine the relationship between point queries within one text instance using an intra-group self-attention across dimension $N$. Here, keys are the same with queries while values only contain the content part: $\textbf{K}_q = \textbf{Q}_q$, $\textbf{V}_q = \textbf{C}_q$. Then, an inter-group self-attention is conducted across $K$ instances to capture the relationship between different instances. The updated composite queries are further sent into the deformable cross-attention to aggregate multi-scale text features from the encoder. The point coordinates $Coords$ are used as the reference points in the deformable attention. 

With the explicit point information flowing in the decoder, we adopt a 3-layer MLP head to predict the offsets and update point coordinates after each decoder layer. Then, the updated coordinates will be transformed into new positional queries by \cref{eq_2}.

\subsection{Task Prediction}
After getting queries of shape $K \times N \times 256$ from decoder for each image, we adopt simple prediction heads to solve sub-tasks. \textbf{(1) Instance classification.} We use a linear projection for binary classification (text or background). During inference, we take the mean of $N$ scores as the confidence score for each instance.
\textbf{(2) Character classification.} As the points are uniformly sampled on the center curve of each text to cover characters, each point query represents a specific class (including background). We adopt a linear projection to perform character classification.
\textbf{(3) Center curve points.} Given the explicit point coordinates $Coords$, a 3-layer MLP head $MLP_{coord}$ is used to predict the coordinate offsets from reference points to ground truth points on the center curve.
\textbf{(4) Boundary points.} Similarly, a 3-layer MLP head $MLP_{bd}$ is used to predict the offsets to the ground truth points on the top and bottom curves. 

\subsection{Optimization}
\noindent \textbf{Bipartite Matching.} After obtaining the prediction set $\hat{Y}$ from the four heads and the ground truth set $Y$, we use the Hungarian algorithm \cite{kuhn1955hungarian} to get an optimal injective function $\varphi: [Y] \mapsto [\hat{Y}]$ that minimizes the matching cost $\mathcal{C}$:
\begin{equation}
    \underset{\varphi}{\arg \min} \sum\limits^{G-1}_{g=0}{\mathcal{C}(Y^{(g)}, \hat{Y}^{(\varphi(g))})}, \label{eq_4}
\end{equation}
where $G$ is the number of ground truth instances per image. 

Regarding the cost $\mathcal{C}$, previous work \cite{zhang2022text} only considers the class and position similarity while ignoring the similarity of the text script. However, matched pairs with similar positions may be quite different in texts, which could increases the optimization difficulty. TTS \cite{kittenplon2022towards} proposes a cross-entropy-based text matching criterion to address this issue, but it is not applicable to our method since the character predictions are not aligned with the ground truth due to background class and repeated characters. We introduce a text matching criterion based on the typical Connectionist Temporal Classification loss \cite{graves2006connectionist} which copes with the length inconsistency issue. For the $g$-th ground truth and its matched query, the complete cost function is:
\begin{equation}
\resizebox{.9\linewidth}{!}{$\begin{aligned}
    \mathcal{C}(Y^{(g)}, \hat{Y}^{(\varphi(g))}) = & \lambda_{\text{cls}}\text{FL}'(\hat{b}^{(\varphi(g))}) + \lambda_{\text{text}} \text{CTC}(t^{(g)}, \hat{t}^{(\varphi(g))}) \\
    & + \lambda_{\text{coord}} \sum_{n=0}^{N-1} \left\| p_n^{(g)} - \hat{p}_n^{(\varphi(g))}\right\|,
\end{aligned}$} \label{eq_5}
\end{equation}
where $\lambda_{\text{cls}}$, $\lambda_{\text{text}}$, and $\lambda_{\text{coord}}$ are hyper-parameters to balance different tasks. $\hat{b}^{(\varphi(g))}$ is the probability for the text-only instance class. The same as \cite{zhang2022text}, $\text{FL}'$ is defined as the difference between the positive and negative term: $\text{FL}'(x)=-\alpha (1-x)^\gamma \log(x) + (1-\alpha) x^\gamma \log(1-x)$, which is derived from the focal loss\cite{lin2017focal}. The second term is the CTC loss between the ground truth text script and predicted $N$ characters. The third term is the L1 distance between the ground truth and predicted point coordinates on the center curve.

\noindent \textbf{Overall Loss.} For the $k$-th query, the focal loss for instance classification is formulated as:
\begin{equation}
\begin{aligned}
    \mathcal{L}_{\text{cls}}^{(k)} = & -\mathds{1}_{\left\{ k \in \text{Im}(\varphi) \right\}} \alpha (1-\hat{b}^{(k)})^\gamma \log (\hat{b}^{(k)}) \\
    & -\mathds{1}_{\left\{ k \notin \text{Im}(\varphi) \right\}} (1-\alpha) (\hat{b}^{(k)})^\gamma \log (1-\hat{b}^{(k)}),
\end{aligned}
\end{equation}
where $\mathds{1}$ is the indicator function, $\text{Im}(\varphi)$ is the image of the mapping $\varphi$. As for character classification, we exploit the CTC loss to address the length inconsistency issue between the ground truth text scripts and predictions:
\begin{equation}
    \mathcal{L}_{\text{text}}^{(k)} = \mathds{1}_{\left\{ k \in \text{Im}(\varphi) \right\}} \text{CTC}(t^{(\varphi^{-1}(k))}, \hat{t}^{(k)}).
\end{equation}
In addition, L1 distance loss is used for supervising points coordinates on the center curve and the boundaries (\ie, the top and bottom curves):
\begin{equation}
    \mathcal{L}_{\text{coord}}^{(k)} = \mathds{1}_{\left\{ k \in \text{Im}(\varphi) \right\}} \sum_{n=0}^{N-1} \left\|p_n^{(\varphi^{-1}(k))} - \hat{p}_n^{(k)} \right\|,
\end{equation}
\begin{equation}
\resizebox{1\linewidth}{!}{$
    \mathcal{L}_{\text{bd}}^{(k)} = \mathds{1}_{\left\{ k \in \text{Im}(\varphi) \right\}} \sum_{n=0}^{N-1} \left(\left\|top_n^{(\varphi^{-1}(k))} - \hat{top}_n^{(k)} \right\| + \left\|bot_n^{(\varphi^{-1}(k))} - \hat{bot}_n^{(k)} \right\|\right).
$}
\end{equation}

The loss function for the decoder consists of the four aforementioned losses:
\begin{equation}
    \mathcal{L}_{\text{dec}} = \sum_{k} \left( \lambda_{\text{cls}} \mathcal{L}_{\text{cls}}^{(k)}  + \lambda_{\text{text}} \mathcal{L}_{\text{text}}^{(k)} + \lambda_{\text{coord}} \mathcal{L}_{\text{coord}}^{(k)} + \lambda_{\text{bd}} \mathcal{L}_{\text{bd}}^{(k)}\right),
\end{equation}
where the hyper-parameters $\lambda_{\text{cls}}$, $\lambda_{\text{text}}$, and $\lambda_{\text{coord}}$ are the same as those in \cref{eq_5}. $\lambda_{\text{bd}}$ is the boundary loss weight. In addition, to make the Bezier center curve proposals in \cref{sec:top-k proposal} more accurate, we resort to adding intermediate supervision on the encoder. As we hope the points sampled on the Bezier center curve proposals are as close to the ground truth as possible, we calculate the L1 loss for the $N$ uniformly sampled points instead of only the four Bezier control points for each instance. This supervision method has been explored by \cite{feng2022rethinking}. The loss function for the encoder is formulated as:
\begin{equation}
    \mathcal{L}_{\text{enc}} = \sum_{i} \left( \lambda_{\text{cls}} \mathcal{L}_{\text{cls}}^{(i)}  + \lambda_{\text{coord}} \mathcal{L}_{\text{coord}}^{(i)}\right),
\end{equation}
where bipartite matching is also exploited to get one-to-one matching. The overall loss $\mathcal{L}$ is defined as:
\begin{equation}
    \mathcal{L} = \mathcal{L}_{\text{dec}} + \mathcal{L}_{\text{enc}}.
\end{equation}

\section{Experiments}
\subsection{Settings}
\noindent \textbf{Benchmarks.} We evaluate our method on Total-Text \cite{ch2020total}, ICDAR 2015 (IC15) \cite{karatzas2015icdar} and CTW1500 \cite{liu2019curved}. \textbf{Total-Text} is an arbitrarily-shaped word-level scene text benchmark, with 1,255 training images and 300 testing images. \textbf{IC15} contains 1,000 training images and 500 testing images for quadrilateral scene text. Different from Total-Text and IC15, \textbf{CTW1500} is a text-line level benchmark for scene text with arbitrary shape. There are 1,000 training images and 500 testing images. We adopt the following additional datasets for pre-training: 1) Synth150K \cite{liu2020abcnet}, a synthetic dataset that contains 94,723 images with multi-oriented texts and 54,327 images with curved texts. 2) ICDAR 2017 MLT (MLT17) \cite{nayef2017icdar2017}, which is a multi-language scene text dataset. 3) ICDAR 2013 (IC13) \cite{karatzas2013icdar} that contains 229 training images with horizontal text. We also investigate the influence of leveraging 21,778 training images and 3,124 validation images of TextOCR \cite{singh2021textocr}.

\noindent \textbf{Implementation Details.} 
The number of heads and sampling points for deformable attention is 8 and 4, respectively. The number of both encoder and decoder layers is 6. The number of proposals $K$ is 100. The number of sampled points $N$ is set to 25 by default without special instructions. Our models predict 37 character classes on Total-Text and IC15, 96 classes on CTW1500. AdamW \cite{loshchilov2017decoupled} is used as the optimizer. The loss weights $\lambda_{\text{cls}}$, $\lambda_{\text{coord}}$,
$\lambda_{\text{bd}}$, and $\lambda_{\text{text}}$ are set to 1.0, 1.0, 0.5, and 0.5, respectively. For focal loss, $\alpha$ is 0.25 and $\gamma$ is 2.0.
The image batch size is 8. More details of the experiments can be found in the appendix. 

\subsection{Ablation Studies}
\label{sec:ablation}
We first conduct ablation experiments on Total-Text. We then investigate the influence of different training datasets and backbone choices. In \cref{tab:ablation_loss} and \cref{tab:ablation_sharing}, we pre-train each model on a mixture of Synth150K and Total-Text, then fine-tune it on Total-Text.

\noindent \textbf{Text Loss Weight.} We study the influence of the text loss weight, which has a direct impact on recognition performance. The results in \cref{tab:ablation_loss} show that our model achieves a better trade-off between detection and end-to-end performance when $\lambda_{\text{text}}$ is set to 0.5. We choose $\lambda_{\text{text}} = 0.5$ for other experiments.

\begin{table}[!h]
    \centering
    \setlength{\tabcolsep}{6pt}
\resizebox{0.8\linewidth}{!}{%
    \begin{tabular}{c|ccc|cc}
    \toprule[1.3pt]
    \multirow{2}{*}{$\lambda_{\text{text}}$} & \multicolumn{3}{c|}{Detection} & \multicolumn{2}{c}{E2E} \\ 
    \cline{2-4} \cline{5-6} 
     & P & R & F1 & None & Full \\ 
    \midrule[1.1pt]
    0.25 & \textbf{94.29} & 82.07 & \textbf{87.76} & 76.68 & 85.76 \\
    \rowcolor{gray!20} \textbf{0.5} & 93.86 & \textbf{82.11} & 87.59 & \textbf{78.83} & \textbf{86.15} \\
    0.75 & 94.15 & 79.27 & 86.07 & 78.82 & 85.71 \\
    1.0 & 93.06 & 81.71 & 87.01 & 77.73 & 85.61 \\
    \bottomrule[1.3pt]
    \end{tabular}%
}
\caption{Hyper-parameter study of $\lambda_{\text{text}}$. E2E: the end-to-end spotting results. None (Full) denotes the F1-measure without (with) using the lexicon that includes all words in the test set.}
\label{tab:ablation_loss}
\end{table}

\noindent \textbf{Sharing Point Embeddings.} In \cref{tab:ablation_sharing}, when sharing point embeddings for all instances, the results on end-to-end task drop. It indicates that different instances require different point embeddings to encode the instance-specific features.

\begin{table}[!t]
    \centering
    \setlength{\tabcolsep}{3pt}
\resizebox{\linewidth}{!}{%
    \begin{tabular}{cc|ccc|cc|c|c}
    \toprule[1.5pt]
    \multirow{2}{*}{Sharing} & \multirow{2}{*}{Matching} & \multicolumn{3}{c|}{Detection} & \multicolumn{2}{c|}{E2E} & \multirow{2}{*}{\#Params} & \multirow{2}{*}{FPS} \\ 
    \cline{3-5} \cline{6-7} 
    && P & R & F1 & None & Full & & \\ 
    \midrule[1.1pt]
    \rowcolor{gray!20} \ding{55} & \checkmark & \textbf{93.86} & \textbf{82.11} & \textbf{87.59} & \textbf{78.83} & \textbf{86.15} &42.5M &17.0 \\
    \checkmark & \checkmark & 93.60 & 81.21 & 86.96 & 77.58 & 85.98 &41.8M &17.0 \\
    \ding{55} & \ding{55} & 92.90 &82.11 &87.17 &77.85 &85.20 &42.5M &17.0\\
    \checkmark & \ding{55} &93.41 &81.98 &87.32 &77.09 &85.13 &41.8M &17.0 \\
    \bottomrule[1.5pt]
    \end{tabular}%
}
\caption{Influence of sharing point embeddings and text matching.}
\label{tab:ablation_sharing}
\end{table}

\begin{table}[!t]
    \centering
    \setlength{\tabcolsep}{2pt}
\resizebox{\linewidth}{!}{%
    \begin{tabular}{c|l|c|ccc|cc}
    \toprule[1.5pt]
    \multirow{2}{*}{\#Row } &\multirow{2}{*}{External Data} &\multirow{2}{*}{Volume} & \multicolumn{3}{c|}{Detection} & \multicolumn{2}{c}{E2E} \\ 
    \cline{4-6} \cline{7-8} 
    & & & P & R & F1 & None & Full \\ 
    \midrule[1.1pt]
    1 & Synth150K & 150K & \textbf{93.86} & 82.11 & 87.59 & 78.83 & 86.15 \\
    2 & Row\#1+MLT17+IC13+IC15 & 160K & 93.09 & 82.11 & 87.26 & 79.65 & 87.00 \\
    \rowcolor{gray!20} 3 & Row\#2 +TextOCR & 185K & 93.19 & \textbf{84.64} & \textbf{88.72} & \textbf{82.54} & \textbf{88.72} \\
    \bottomrule[1.5pt]
    \end{tabular}%
}
\caption{Influence of the training data. Volume: dataset volume.}
\label{tab:ablation_data}
\vspace{-3mm}
\end{table}

\noindent \textbf{Text Matching Criterion.} To evaluate the effectiveness of the text matching criterion, in \cref{tab:ablation_sharing}, we remove the text matching criterion from \cref{eq_5}. The primary end-to-end results decline. It validates the value of conducting text matching, which provides high-quality one-to-one matching between the predictions and ground truth according to both the position and script similarity.

\noindent \textbf{Training Data.} We study the influence of different pre-training data in \cref{tab:ablation_data}. For the end-to-end spotting task, with only Synth150K as the pre-training data, our method can achieve 78.83\% accuracy without using lexicon. With additional MLT17, IC13, and IC15 real training data, the `None' and `Full' scores are improved by 0.82\% and 0.85\%, respectively. We further show that the performance is improved by a large margin using TextOCR. In TextOCR, the average number of text instances per image is higher than in other datasets, which can provide more positive signals. It demonstrates the value of using real data for pre-training and the scalability of our model on different data.

\begin{figure}[!t]
    \centering
    \includegraphics[width=1.0\linewidth]{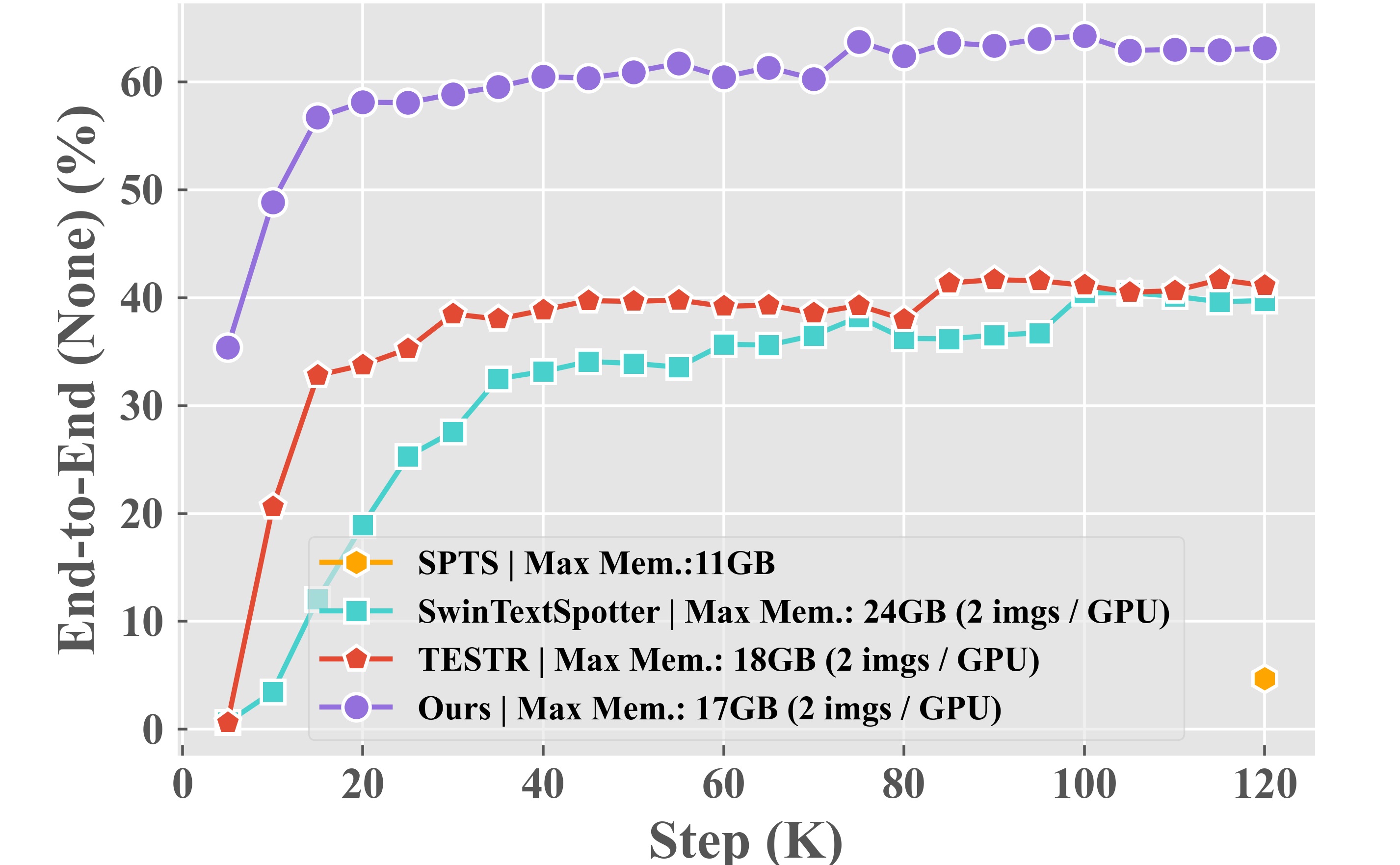}
    \caption{Comparison with open-source Transformer-based methods using only Total-Text training set.}
    \label{fig:train_eff}
\end{figure}

We further compare DeepSolo with existing open-source Transformer-based methods \cite{huang2022swintextspotter,zhang2022text,peng2022spts} by only using the training set of Total-Text. For a fair comparison, we apply the same data augmentation. Since both TESTR and our method are based on Deformable-DETR \cite{zhu2020deformable}, we set the same configuration for the Transformer modules. ResNet-50 \cite{he2016deep} is adopted in all experiments. The image batch size is set to 8, while the actual batch size of SPTS doubles due to batch augmentation. In \cref{fig:train_eff}, our method achieves faster convergence and better performance, showing the superiority of DeepSolo over representative ones in training efficiency. 
In addition, the training GPU memory usage of our model is also less than SwinTextSpotter and TESTR.

Compared with TESTR which adopts dual decoders, the results demonstrate the value of our proposed query form, which contributes a unified representation to detection and recognition, encodes a more accurate and explicit position prior, and facilitates the learning of the single decoder. Consequently, instead of using complex dual decoders, our simpler design could effectively mitigate the synergy issue.

\noindent \textbf{Different Backbones.} We also conduct experiments to investigate the influence of different backbones on our model in \cref{tab:ablation_backbone}. We select ResNet, Swin Transformer \cite{liu2021swin}, and ViTAE-v2 \cite{zhang2022vitaev2} for comparison. All the models are pre-trained with the mixture data as listed in Row \#2 of \cref{tab:ablation_data}. Compared with ResNet-50, ViTAE-v2-S outperforms it by a large margin on the end-to-end spotting task, \ie, 2.14 \% on the `None' setting. Compared with ResNet-101, Swin-S achieves a gain of 1.15\% on the ‘None’ setting. We conjecture that the domain gap between large-scale synthetic data and small-scale real data might impact the performance of different backbones. It deserves further exploration to carefully tune the training configuration and study the training paradigm for text spotting.

\begin{table}[!t]
    \centering
    \setlength{\tabcolsep}{4pt}
\resizebox{\linewidth}{!}{%
    \begin{tabular}{c|ccc|cc|c|c}
    \toprule[1.5pt]
    \multirow{2}{*}{Backbone} & \multicolumn{3}{c|}{Detection} & \multicolumn{2}{c|}{E2E} & \multirow{2}{*}{\#Params} & \multirow{2}{*}{\makecell[c]{Mem.\\(MB)}} \\ 
    \cline{2-4} \cline{5-6} 
     & P & R & F1 & None & Full & & \\ 
    \midrule[1.1pt]
    ResNet-50 &93.09 &82.11 &87.26 &79.65 &87.00 &42.5M &17,216 \\
    Swin-T &92.77 &83.51 &87.90 &79.66 &87.05 &43.1M &26,573 \\
    ViTAEv2-S &92.57 &\textbf{85.50} &\textbf{88.89} &\textbf{81.79} &\textbf{88.40} &33.7M &25,332 \\
    ResNet-101 &93.20 &83.51 &88.09 &80.12 &87.14 &61.5M &19,541 \\
    Swin-S &\textbf{93.72} &84.24 &88.73 &81.27 &87.75 &64.4M &33,974 \\
    \bottomrule[1.5pt]
    \end{tabular}%
}
\caption{The influence of different backbones. Mem.: the peak memory of batching two images on one GPU during pre-training.}
\label{tab:ablation_backbone}
\vspace{-3mm}
\end{table}

\begin{table*}[!t]
\centering
\setlength{\tabcolsep}{5pt}
\resizebox{\linewidth}{!}{
\begin{tabular}{l|l|ccc|cc|c|c}
\toprule[1.5pt]
\multirow{2}{*}{Method} & \multirow{2}{*}{External Data} & \multicolumn{3}{c|}{Detection} & \multirow{2}{*}{None} & \multirow{2}{*}{Full} & \multirow{2}{*}{\makecell[c]{FPS\\(report)}} & \multirow{2}{*}{\makecell[c]{FPS\\(A100)}} \\
\cline{3-5}
& &P &R &F1 & & & & \\
\midrule[1.1pt]
TextDragon \cite{feng2019textdragon} &Synth800K &85.6 &75.7 &80.3 &48.8 &74.8 &$-$ &$-$ \\
CharNet \cite{xing2019convolutional} $^{*}$ &Synth800K &88.6 &81.0 &84.6 &63.6 &$-$ &$-$ &$-$ \\
TextPerceptron \cite{qiao2020text} &Synth800K &88.8 &81.8 &85.2 &69.7 &78.3 &$-$ &$-$\\
CRAFTS \cite{baek2020character} $^{*}$ &Synth800K+IC13 &89.5 &85.4 &87.4 &78.7 &$-$ &$-$ &$-$ \\
Boundary \cite{wang2020all} &Synth800K+IC13+IC15 &88.9 &85.0 &87.0 &65.0 &76.1 &$-$ &$-$ \\
Mask TextSpotter v3 \cite{liao2020mask} &Synth800K+IC13+IC15+SCUT &$-$ &$-$ &$-$ &71.2 &78.4 &$-$ &$-$ \\
PGNet \cite{wang2021pgnet} &Synth800K+IC15 &85.5 &86.8 &86.1 &63.1 &$-$ &35.5 &$-$ \\
MANGO \cite{qiao2021mango} $^{*}$ &Synth800K+Synth150K+COCO-Text+MLT19+IC13+IC15 &$-$ &$-$ &$-$ &72.9 &83.6 &4.3 &$-$ \\
PAN++ \cite{wang2021pan++} &Synth800K+COCO-Text+MLT17+IC15 &$-$ &$-$ &$-$ &68.6 &78.6 &21.1 &$-$ \\
ABCNet v2 \cite{liu2021abcnet} &Synth150K+MLT17 &90.2 &84.1 &87.0 &70.4 &78.1 &10.0 &14.9 \\
SRSTS \cite{wu2022decoupling} &Synth800K+Synth150K+COCO-Text+MLT17+ArT19+IC15 &92.0 &83.0 &87.2 &78.8 &86.3 &18.7 &$-$\\
GLASS \cite{ronen2022glass} &Synth800K &90.8 &85.5 &88.1 &79.9 &86.2 &3.0 &$-$ \\
\midrule
TESTR \cite{zhang2022text} &Synth150K+MLT17 &93.4 &81.4 &86.9 &73.3 &83.9 &5.3 &12.1 \\
SwinTextSpotter \cite{huang2022swintextspotter} &Synth150K+MLT17+IC13+IC15 &$-$ &$-$ &88.0 &74.3 &84.1 &$-$ &2.9 \\
SPTS \cite{peng2022spts} &Synth150K+MLT17+IC13+IC15 &$-$ &$-$ &$-$ &74.2 &82.4 &$-$ &0.6 \\
TTS (poly) \cite{kittenplon2022towards} &Synth800K+COCO-Text+IC13+IC15+SCUT &$-$ &$-$ &$-$ &78.2 &86.3 &$-$ &$-$ \\
\rowcolor{gray!20} DeepSolo (ResNet-50) &Synth150K &93.9 & 82.1 & 87.6 & 78.8 & 86.2 &17.0 &17.0 \\
\rowcolor{gray!20} DeepSolo (ResNet-50) &Synth150K+MLT17+IC13+IC15 &93.1 & 82.1 & 87.3 & 79.7 & 87.0 &17.0 &17.0 \\
\rowcolor{gray!20} DeepSolo (ResNet-50) &Synth150K+MLT17+IC13+IC15+TextOCR &93.2 & 84.6 & \underline{88.7} & \underline{82.5} & \underline{88.7} &17.0 &17.0 \\
\rowcolor{gray!20} DeepSolo (ViTAEv2-S) &Synth150K+MLT17+IC13+IC15+TextOCR &92.9 & 87.4 & \textbf{90.0} & \textbf{83.6} & \textbf{89.6} &10.0 &10.0 \\
\bottomrule[1.5pt]\end{tabular}}
\caption{Performance on Total-Text. ‘*’: character-level annotations are used. ‘*’ has the same meaning for other tables.}
\label{tab:main_totaltext}
\end{table*}

\begin{figure*}[!ht]
    \centering
    \includegraphics[width=\linewidth]{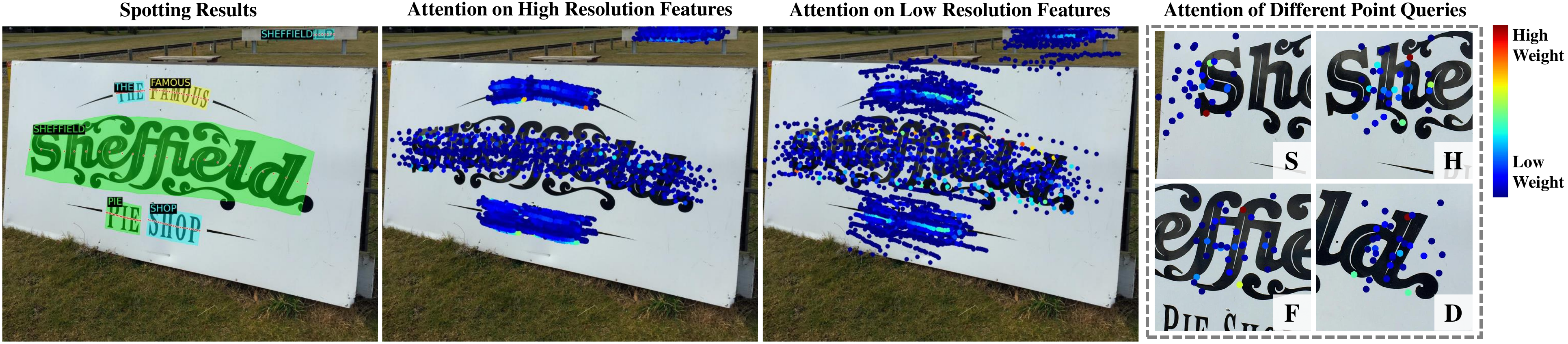}
    \caption{Visualization of the spotting results, attention on different scale features, and attention of point queries.}
    \label{fig:attn_vis}
\end{figure*}

\subsection{Comparison with State-of-the-art Methods}
\noindent \textbf{Results on Total-Text.} To evaluate the effectiveness of DeepSolo on scene text with arbitrary shape, we compare our model with state-of-the-art methods in \cref{tab:main_totaltext}. The metrics in the end-to-end setting are the primary ones for text spotting. \textbf{1)} Considering the `None' results, with only Synth150K as the external data, our method surpasses previous methods except GLASS. Compared with other Transformer-based methods, DeepSolo significantly outperforms TESTR, SwinTextSpotter, and SPTS by 5.5\%, 4.5\%, and 4.6\%, respectively. DeepSolo also outperforms TTS by 0.6\% while using far less training data. \textbf{2)} With additional MLT17, IC13, and IC15 real data, DeepSolo achieves 79.7\% in the `None' setting, which is comparable with the 79.9\% performance of GLASS. Note that DeepSolo runs faster than GLASS and there is no elaborately tailored module for recognition, \eg, the Global to Local Attention Feature Fusion and the external recognizer in GLASS, while we only use a simple linear layer for recognition output. \textbf{3)} When using TextOCR, our method achieves very promising spotting performance, \ie, 82.5\% and 88.7\% at the `None' and `Full' settings. With ViTAEv2-S, the results are further promoted.

\noindent \textbf{Results on ICDAR 2015.} We conduct experiments on ICDAR 2015 to verify the effectiveness of DeepSolo on multi-oriented scene text, as presented in \cref{tab:main_ic15}. The results show that DeepSolo achieves decent performance among the comparing methods. Specifically, compared with Transformer-based methods using the same training datasets, DeepSolo surpasses SwinTextSpotter and SPTS by 6.4\% and 11.1\% on the E2E setting with the generic lexicon. With TextOCR, DeepSolo (ResNet-50) achieves the ‘S’, ‘W’, and ‘G’ metrics of 88.0\%, 83.5\%, and 79.1\%.

\noindent \textbf{Results on CTW1500.} In \cref{tab:main_ctw}, pre-trained with Synth150K, DeepSolo with the maximum recognition length of 25 already outperforms most of the previous approaches on the ‘None’ metric. We further increase the number of point queries from 25 to 50 for each text instance, achieving absolute 3.1$\%$ improvement on the ‘None’ result, without obvious sacrifice on inference speed. With MLT17, IC13, IC15 and Total-Text as the additional datasets, DeepSolo presents 64.2$\%$ performance without the help of lexicons, being 0.6$\%$ better and 25 times faster than SPTS.

\begin{figure}[!t]
    \centering
    \includegraphics[width=1\linewidth]{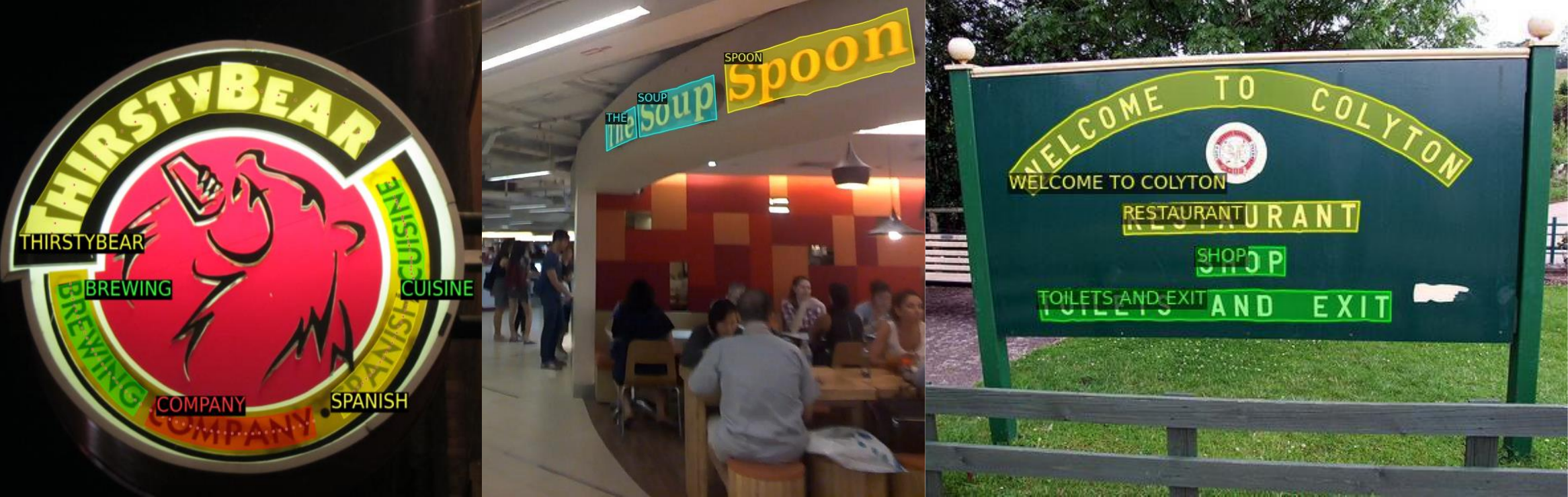}
    \caption{Visualizations on Total-Text, IC15 and CTW1500.}
    \label{fig:qualitative_result}
\end{figure}

\begin{table*}[!t]
\centering
\setlength{\tabcolsep}{5pt}
\resizebox{\linewidth}{!}{
\begin{tabular}{l|l|ccc|ccc|ccc}
\toprule[1.5pt]
\multirow{2}{*}{Method} & \multirow{2}{*}{External Data} & \multicolumn{3}{c|}{Detection} & \multicolumn{3}{c|}{E2E} & \multicolumn{3}{c}{Word Spotting} \\
\cline{3-5} \cline{6-8} \cline{9-11} 
& &P &R &F1 &S &W &G &S &W &G \\
\midrule[1.1pt]
TextDragon \cite{feng2019textdragon} &Synth800K &92.5 &83.8 &87.9 &82.5 &78.3 &65.2 &86.2 &81.6 &68.0 \\
CharNet \cite{xing2019convolutional} $^{*}$ &Synth800K &91.2 &88.3 &89.7 &80.1 &74.5 &62.2 &$-$ &$-$ &$-$ \\
TextPerceptron \cite{qiao2020text} &Synth800K &92.3 &82.5 &87.1 &80.5 &76.6 &65.1 &84.1 &79.4 &67.9 \\
CRAFTS \cite{baek2020character} $^{*}$ &Synth800K+IC13 &89.0 &85.3 &87.1 &83.1 &82.1 &74.9 &$-$ &$-$ &$-$ \\
Boundary \cite{wang2020all} &Synth800K+IC13+Total-Text &89.8 &87.5 &88.6 &79.7 &75.2 &64.1 &$-$ &$-$ &$-$ \\
Mask TextSpotter v3 \cite{liao2020mask} &Synth800K+IC13+Total-Text+SCUT &$-$ &$-$ &$-$ &83.3 &78.1 &74.2 &83.1 &79.1 &75.1 \\
PGNet \cite{wang2021pgnet} &Synth800K+Total-Text &91.8 &84.8 &88.2 &83.3 &78.3 &63.5 &$-$ &$-$ &$-$ \\
MANGO \cite{qiao2021mango} $^{*}$ &Synth800K+Synth150K+COCO-Text+MLT19+IC13+Total-Text &$-$ &$-$ &$-$ &85.4 &80.1 &73.9 &85.2 &81.1 &74.6 \\
PAN++ \cite{wang2021pan++} &Synth800K+COCO-Text+MLT17+Total-Text &$-$ &$-$ &$-$ &82.7 &78.2 &69.2 &$-$ &$-$ &$-$ \\
ABCNet v2 \cite{liu2021abcnet} &Synth150K+MLT17 &90.4 &86.0 &88.1 &82.7 &78.5 &73.0 &$-$ &$-$ &$-$ \\
SRSTS \cite{wu2022decoupling} &Synth800K+Synth150K+COCO-Text+MLT17 &96.1 &82.0 &88.4 &85.6 &81.7 &74.5 &85.8 &82.6 &76.8 \\
GLASS \cite{ronen2022glass} &Synth800K &86.9 &84.5 &85.7 &84.7 &80.1 &76.3 &86.8 &82.5 &78.8 \\
\midrule
TESTR \cite{zhang2022text} &Synth150K+MLT17+Total-Text &90.3 &89.7 &\underline{90.0} &85.2 &79.4 &73.6 &$-$ &$-$ &$-$ \\
SwinTextSpotter \cite{huang2022swintextspotter} &Synth150K+MLT17+IC13+Total-Text &$-$ &$-$ &$-$ &83.9 &77.3 &70.5 &$-$ &$-$ &$-$ \\
SPTS \cite{peng2022spts} &Synth150K+MLT17+IC13+Total-Text &$-$ &$-$ &$-$ &77.5 &70.2 &65.8 &$-$ &$-$ &$-$ \\
TTS \cite{kittenplon2022towards} &Synth800K+COCO-Text+IC13+Total-Text+SCUT &$-$ &$-$ &$-$ &85.2 &81.7 &77.4 &85.0 &81.5 &77.3 \\
\rowcolor{gray!20} DeepSolo (ResNet-50) &Synth150K+MLT17+IC13+Total-Text &92.8 &87.4 &\underline{90.0} &86.8 &81.9 &76.9 &86.3 &82.3 &77.3 \\
\rowcolor{gray!20} DeepSolo (ResNet-50) &Synth150K+MLT17+IC13+Total-Text+TextOCR &92.5 &87.2 &89.8 &\underline{88.0} &\underline{83.5} &\underline{79.1} &\underline{87.3} &\underline{83.8} &\underline{79.5} \\
\rowcolor{gray!20} DeepSolo (ViTAEv2-S) &Synth150K+MLT17+IC13+Total-Text+TextOCR &92.4 &87.9 &\textbf{90.1} &\textbf{88.1} &\textbf{83.9} &\textbf{79.5} &\textbf{87.8} &\textbf{84.5} &\textbf{80.0} \\
\bottomrule[1.5pt]\end{tabular}}
\caption{Performance on ICDAR2015. ‘S’, ‘W’ and ‘G’ refer to using strong, weak and generic lexicons.}
\label{tab:main_ic15}
\end{table*}
\begin{table}[!t]
\centering
\resizebox{0.9\linewidth}{!}{
\begin{tabular}{l|cc|c}
\toprule[1.5pt]
Method & None & Full & FPS \\
\midrule[1.1pt]
ABCNet v2 (100 length) \cite{liu2021abcnet} &57.5 &77.2 &10.0 \\
MANGO (25 length)\cite{qiao2021mango} &58.9 &78.7 &8.4 \\
\midrule
SwinTextSpotter \cite{huang2022swintextspotter} &51.8 &77.0 &$-$ \\
TESTR (100 length) \cite{zhang2022text} &56.0 &\underline{81.5} &15.9 \dag \\
SPTS (100 length) \cite{peng2022spts} &\underline{63.6} &\textbf{83.8} &0.8 \dag \\
\rowcolor{gray!20} DeepSolo (25 length, Synth150K) &60.1 &78.4 &20.0 \dag \\
\rowcolor{gray!20} DeepSolo (50 length, Synth150K) &63.2 &80.0 &20.0 \dag \\
\rowcolor{gray!20} DeepSolo (50 length) &\textbf{64.2} &81.4 &20.0 \dag \\
\bottomrule[1.5pt]\end{tabular}}
\caption{End-to-end recognition performance on CTW1500. ResNet-50 is adopted in DeepSolo. ‘length’: the maximum recognition length. ‘\dag’: the FPS measured on one A100 GPU.}
\label{tab:main_ctw}
\vspace{-3mm}
\end{table}

\subsection{Visual Analysis}
\cref{fig:attn_vis} visualizes the spotting results, the attention on different scale features, and the attention of different point queries. It shows that DeepSolo is capable of correctly recognizing scene texts of large size variance. 
In the rightest figure, it is noteworthy that point queries highly attend to the discriminative extremities of characters, which indicates that point queries can effectively encode the character position, scale, and semantic information. More qualitative results are presented in \cref{fig:qualitative_result}.

\subsection{Compatibility to Line Annotations}
\begin{figure}[!t]
    \centering
    \includegraphics[width=1\linewidth]{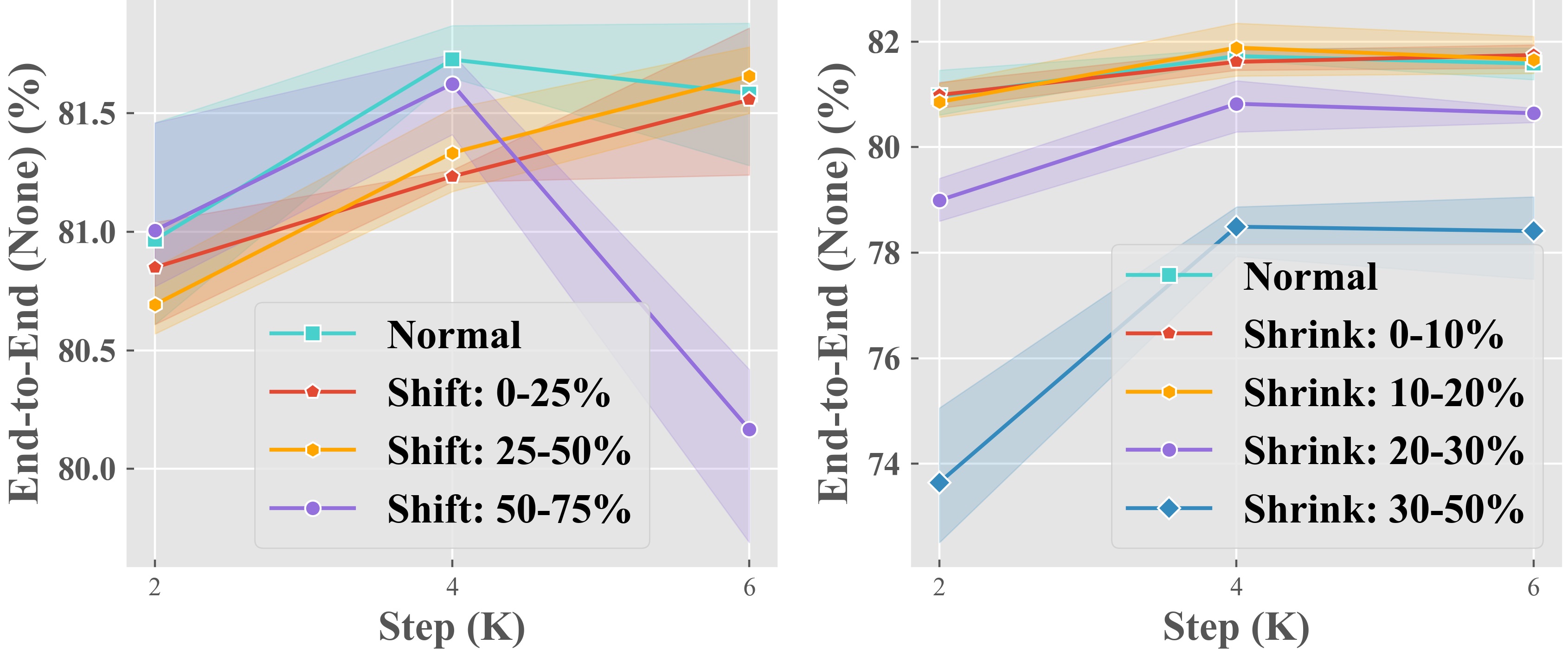}
    \caption{Analysis of the sensitivity to different line locations.}
    \label{fig:noise_analysis}
\end{figure}

DeepSolo can not only adapt to polygon annotations but also line annotations, which are much easier to obtain. We conduct experiments on Total-Text as the target dataset with only line annotations. To take the advantage of existing full annotations, we first pre-train the model on a mixture of Synth150K, MLT17, IC13, IC15, and TextOCR. Then, we simply exclude the boundary head and fine-tune the model on Total-Text with IC13 and IC15 for 6K steps, using only the text center line annotations. During fine-tuning, the random crop augmentation which needs box information is discarded. We use the evaluation protocol provided by SPTS \cite{peng2022spts}. To further study the sensitivity to the line location, we randomly shift the center line annotations to the boundary and shrink them to the center point at different levels to simulate annotation errors. Results are plotted in \cref{fig:noise_analysis}. It can achieve 81.6\% end-to-end (‘None') performance, which is comparable with the fully supervised model, \ie, 82.5\%. As can be seen in \cref{fig:noise_analysis}, the model is robust to the shift from 0\% to 50\% and the shrinkage from 0\% to 20\%. It indicates that the center line should not be too close to the text boundaries and better cover the complete character area. 

\section{Conclusion}
In this paper, we propose DeepSolo, a succinct DETR-like baseline, taking the merit of our novel explicit point query form that provides a joint and pivotal representation for detection and recognition. With a single decoder and several simple prediction heads, we present a much simpler text spotter compared with previous related methods. Experiments demonstrate DeepSolo has achieved state-of-the-art performance, high training efficiency, and compatibility with different annotations. We hope DeepSolo can serve as a strong baseline and inspire more follow-up works.

\noindent \textbf{Acknowledgements}. This work was supported in part by the National Natural Science Foundation of China under Grants 62076186 and 62225113, and in part by the Science and Technology Major Project of Hubei Province (Next-Generation AI Technologies) under Grant 2019AEA170. Dr. Jing Zhang is supported by the ARC FL-170100117. The numerical calculations have been done on the system in the Supercomputing Center of Wuhan University.

\appendix
{\centering\section*{Appendix}}
\section{Performance on ICDAR 2013}
On ICDAR 2013 (IC13) benchmark for horizontal scene text, DeepSolo is pre-trained on a mixture of Synth150K, MLT17, Total-Text, IC13, IC15, then fine-tuned on IC13 for 1K iterations. During evaluation, we resize the shorter sizes of images to 1,000 while keeping the longer ones shorter than 1,824 pixels. \cref{tab:ic13} compares the performance of our method with previous models. DeepSolo achieves the performance of 95.1$\%$, 93.7$\%$, and 90.1$\%$ on three metrics.

\begin{table}[h]
    \centering
    \setlength{\tabcolsep}{8pt}
\resizebox{0.8\linewidth}{!}{%
    \begin{tabular}{l|ccc}
    \toprule[1.3pt]
    \multirow{2}{*}{Method} & \multicolumn{3}{c}{E2E} \\
    \cline{2-4} 
    &S &W &G \\ 
    \midrule
    MaskTextSpotter \cite{lyu2018mask} &92.2 &91.1 &86.5 \\
    MaskTextSpotter v2 \cite{liao2021mask} &93.3 &91.3 &88.2 \\
    MANGO \cite{qiao2021mango} &\underline{93.4} &\underline{92.3} &\underline{88.7} \\
    SPTS \cite{peng2022spts} &93.3 &91.7 &88.5 \\
    \rowcolor{gray!20} DeepSolo (ResNet-50) &\textbf{95.1} &\textbf{93.7} &\textbf{90.1} \\
    \bottomrule[1.3pt]
    \end{tabular}}
\caption{End-to-end text spotting results on ICDAR 2013.}
\label{tab:ic13}
\vspace{-4mm}
\end{table}

\begin{figure}[h]
    \centering
    \includegraphics[width=\linewidth]{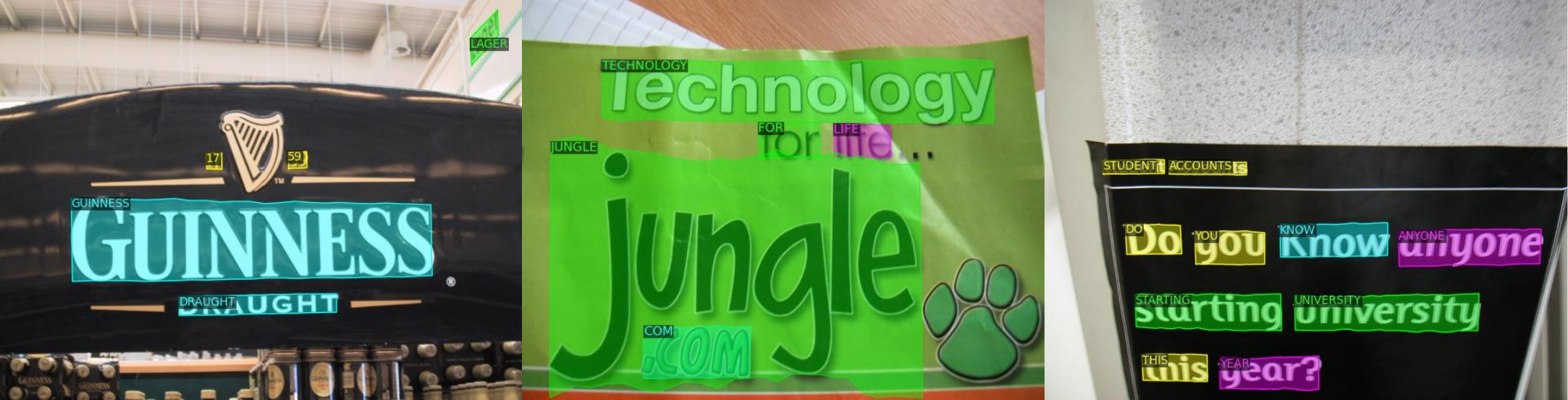}
    \caption{Qualitative results on ICDAR 2013.}
    \label{fig:ic13}
    \vspace{-5mm}
\end{figure}

\begin{table}[!t]
    \centering
    \setlength{\tabcolsep}{8pt}
\resizebox{\linewidth}{!}{%
    \begin{tabular}{l|cc}
    \toprule[1.3pt]
    \multirow{2}{*}{Method} & \multicolumn{2}{c}{E2E} \\
    \cline{2-3} 
    &None &Full \\ 
    \midrule
    MaskTextSpotter v2 \cite{liao2021mask} &39.0 &43.5 \\
    ABCNet \cite{liu2020abcnet} &22.2 &34.3 \\
    ABCNet v2 \cite{liu2021abcnet} &34.5 &47.4 \\
    TESTR \cite{zhang2022text} &34.2 &41.6 \\
    SwinTextSpotter \cite{huang2022swintextspotter} &55.4 &67.9 \\
    SPTS \cite{peng2022spts} &38.3 &46.2 \\
    \rowcolor{gray!20} DeepSolo (ResNet-50, data-1) & 47.6 & 53.0 \\
    \rowcolor{gray!20} DeepSolo (ResNet-50, data-2) &48.5(\textcolor{blue}{+0.9}) &53.9(\textcolor{blue}{+0.9}) \\
    \rowcolor{gray!20} DeepSolo (ResNet-50, data-3) &\underline{64.6}(\textcolor{blue}{+17.0}) &71.2(\textcolor{blue}{+18.2}) \\
    \rowcolor{gray!20} DeepSolo (ViTAEv2-S, data-3) &\textbf{68.8}(\textcolor{blue}{+21.2}) &\underline{75.8}(\textcolor{blue}{+22.8}) \\
    \midrule
    ABCNet v2 w/ Pos.Label \ddag &62.2 &\textbf{76.7} \\
    TESTR w/ Pos.Label \ddag &63.1 &75.4 \\
    SwinTextSpotter \ddag &62.9 &74.7 \\
    \bottomrule[1.3pt]
    \end{tabular}}
\caption{End-to-end text spotting results on Inverse-Text. ‘data-1' denotes the external dataset is Synth150K as in Tab. 5 of the main paper. ‘data-2': ‘Synth150K+MLT17+IC13+IC15'. ‘data-3': ‘Synth150K+MLT17+IC13+IC15+TextOCR'. ‘\ddag': results from \cite{ye2022dptext}, using extensive rotation augmentation.}
\label{tab:inverse-text}
\vspace{-4mm}
\end{table}

\begin{figure}[!t]
    \centering
    \includegraphics[width=\linewidth]{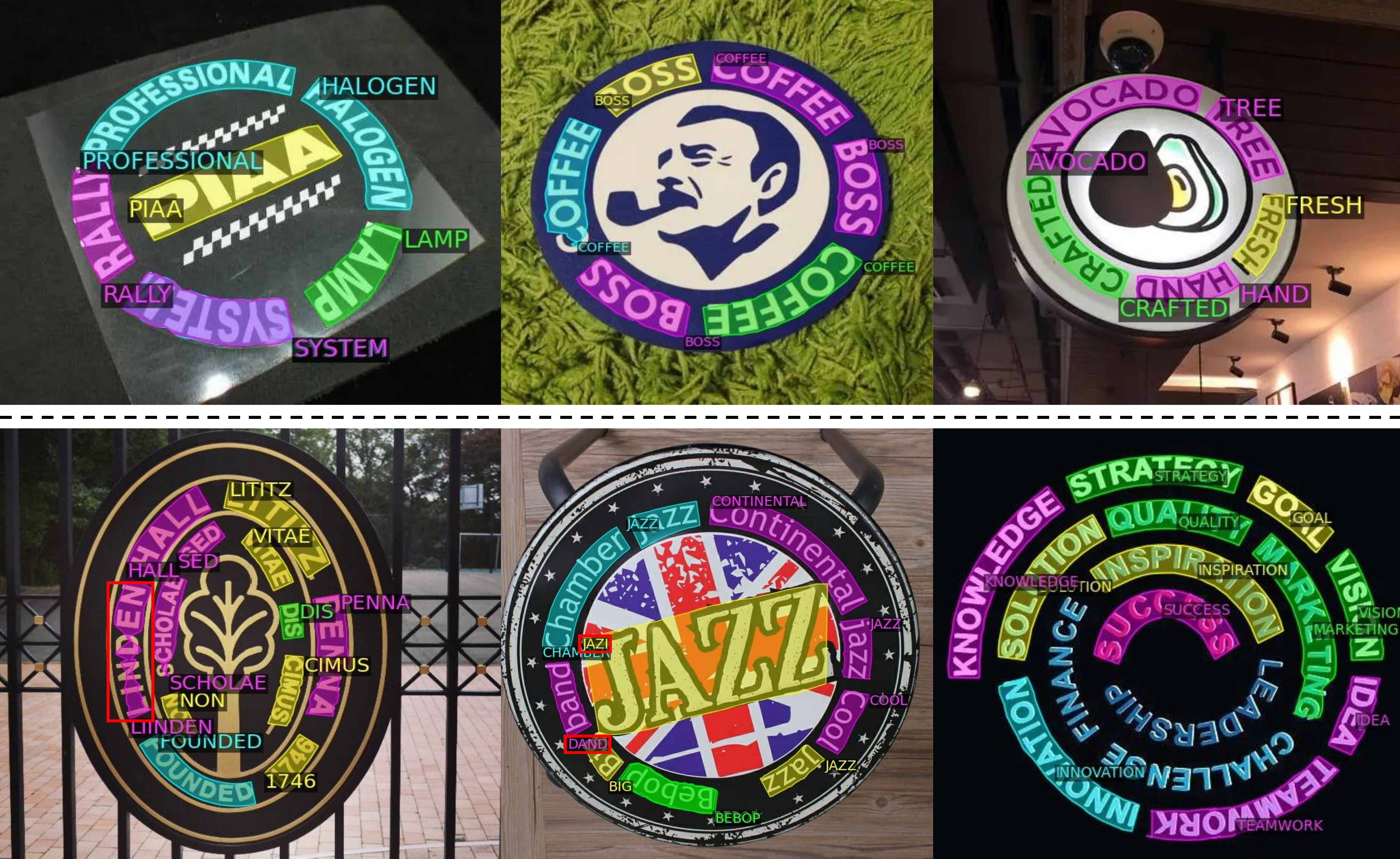}
    \caption{Qualitative results on Inverse-Text. Some failure cases are shown in the bottom row.}
    \label{fig:inverse-text}
    \vspace{-5mm}
\end{figure}

\section{Performance on Inverse-Text}
Inverse-Text \cite{ye2022dptext} is a newly proposed test set for verifying the robustness of scene text detectors and spotters on highly rotated texts. It consists of 500 testing images for arbitrary-shape scene text, with about 40$\%$ inverse-like instances. The DeepSolo models reported in Tab. 5 of the main paper are directly evaluated on Inverse-Text. Results are shown in \cref{tab:inverse-text}. Note that we do not use a stronger rotation augmentation, \eg, angle chosen from $[-90^{\circ}, +90^{\circ}]$ as in SwinTextSpotter. Surprisingly, when TextOCR is leveraged, the ‘None' and ‘Full' performance are additionally improved by $16.1\%$ and $17.3\%$, respectively. While replacing ResNet-50 with ViTAEv2-S, there are $4.2\%$ and $4.6\%$ absolute gain on the two metrics.

\begin{table*}[!t]
\centering
\setlength{\tabcolsep}{2pt}
\resizebox{\linewidth}{!}{
\begin{tabular}{c|l|lccc|lccc|l}
\toprule
\multirow{2}{*}{\#Row} & \multirow{2}{*}{Backbone} & \multicolumn{4}{c|}{Pre-training} & \multicolumn{4}{c|}{Fine-tuning} & \multirow{2}{*}{Where in the Main Paper} \\
\cline{3-6} \cline{7-10}
& &Training Data &$lr$ (Backbone) &Iterations &Step &Training Data &$lr$ (Backbone) &Iterations &Step & \\
\midrule
1 &\multirow{7}{*}{ResNet-50} &Synth150K+Total-Text &$1e^{-4}$ ($1e^{-5}$) &350K &300K &Total-Text &$1e^{-5}$ ($1e^{-6}$) &10K &$-$ &Tab. 1, 2, 3, 5\\
2 & &Synth150K+Total-Text+MLT17+IC13+IC15 &$1e^{-4}$ ($1e^{-5}$) &375K &320K &Total-Text &$1e^{-5}$ ($1e^{-6}$) &10K &$-$ &Tab. 3, 4, 5\\
3 & &Synth150K+Total-Text+MLT17+IC13+IC15+TextOCR &$1e^{-4}$ ($1e^{-5}$) &435K &375K &Total-Text &$1e^{-5}$ ($1e^{-6}$) &2K &$-$ &Tab. 3, 5\\
4 & &Synth150K+Total-Text+MLT17+IC13+IC15 &$1e^{-4}$ ($1e^{-5}$) &375K &320K &IC15 &$1e^{-5}$ ($1e^{-6}$) &3K &$-$ &Tab. 6\\
5 & &Synth150K+Total-Text+MLT17+IC13+IC15+TextOCR &$1e^{-4}$ ($1e^{-5}$) &435K &375K &IC15 &$1e^{-5}$ ($1e^{-6}$) &1K &$-$ &Tab. 6 \\
6 & &Total-Text &$1e^{-4}$ ($1e^{-5}$) &120K &80K &$-$ &$-$ &$-$ &$-$ &Fig. 3\\
7 & &Synth150K+MLT17+IC13+IC15+TextOCR &$1e^{-4}$ ($1e^{-5}$) &435K &375K &Total-Text+IC13+IC15 &$2e^{-5}$ ($2e^{-6}$) &6K &$-$ &Fig. 6\\
8 & &Synth150K+Total-Text+MLT17+IC13+IC15 &$1e^{-4}$ ($1e^{-5}$) &375K &320K &CTW1500 &$5e^{-5}$ ($5e^{-6}$) &12K &8K &Tab. 7\\
\midrule
9 &\multirow{1}{*}{ResNet-101} &Synth150K+Total-Text+MLT17+IC13+IC15 &$1e^{-4}$ ($1e^{-5}$) &375K &320K &Total-Text &$1e^{-5}$ ($1e^{-6}$) &10K &$-$ &Tab. 4\\
\bottomrule
\end{tabular}
}
\caption{Training details of DeepSolo with ResNet. ``Step'' denotes the iteration step where the learning rate is divided by 10.}
\label{tab:resnet}
\end{table*}

\noindent \textbf{Visual Analysis.} Some qualitative results from DeepSolo (ViTAEv2-S) are illustrated in \cref{fig:inverse-text}. In the first row, some inverse-like instances can be correctly recognized. However, as shown in the bottom row, some boundary predictions are not stable, resulting in invalid polygons. Adding boundary points matching may be helpful. Besides, some detection results of inverse instances are filtered because of low confidence score, which can be simply improved by adopting more extensive rotation augmentation.

\section{More Details}
\label{sec:more_details}

The data augmentations include: 1) random rotation with angle chosen from $[-45^{\circ}, +45^{\circ}]$; 2) instance-aware random crop; 3) random resize and 4) color jitter. For inference on Total-Text and CTW1500, the image shorter sides are resized to 1,000. For IC15, following \cite{liao2020mask,zhang2022text}, the shorter sizes are resized to 1,440.

\subsection{Details of DeepSolo with ResNet}
In this subsection, the training details of DeepSolo with ResNet \cite{he2016deep} (ImageNet pre-trained weights from TORCHVISION) are listed in \cref{tab:resnet} with corresponding training data. In Fig. 3 of the main paper, the training schedule of DeepSolo is related to Row \#6, \ie, only the Total-Text training set is utilized. For SPTS \cite{peng2022spts}, we only plot the final performance since it needs much more data and a longer training schedule to achieve ideal performance. The training setting of DeepSolo with line labels is provided in Row \#7. Note that during fine-tuning, the line annotations are used and stronger rotation augmentation (angle randomly chosen from $[-90^{\circ}, +90^{\circ}]$) is adopted.

\subsection{Details of DeepSolo with Swin Transformer}
\label{sec:details_swin}
In Tab. 4 of the main paper, with Swin Transformer \cite{liu2021swin}, we pre-train the model for 375K iterations and fine-tune it on Total-Text for 10K iterations. No part of the backbone is frozen. During pre-training, the initial learning rate for the backbone is $1e^{-4}$. The drop path rate of Swin-T and Swin-S is set to 0.2 and 0.3, respectively. During fine-tuning, we set the learning rate for the backbone to $1e^{-5}$, and the drop path rate to 0.2 and 0.3 for Swin-T and Swin-S. Other training schedules are the same as Row \#2 in \cref{tab:resnet}.

\subsection{Details of DeepSolo with ViTAE}
With ViTAE-v2-S \cite{zhang2022vitaev2}, the drop path rate is set to 0.3 for pre-training and 0.2 for fine-tuning. Other schedules are the same as Swin-S backbone.

\begin{figure}[!t]
    \centering
    \includegraphics[width=\linewidth]{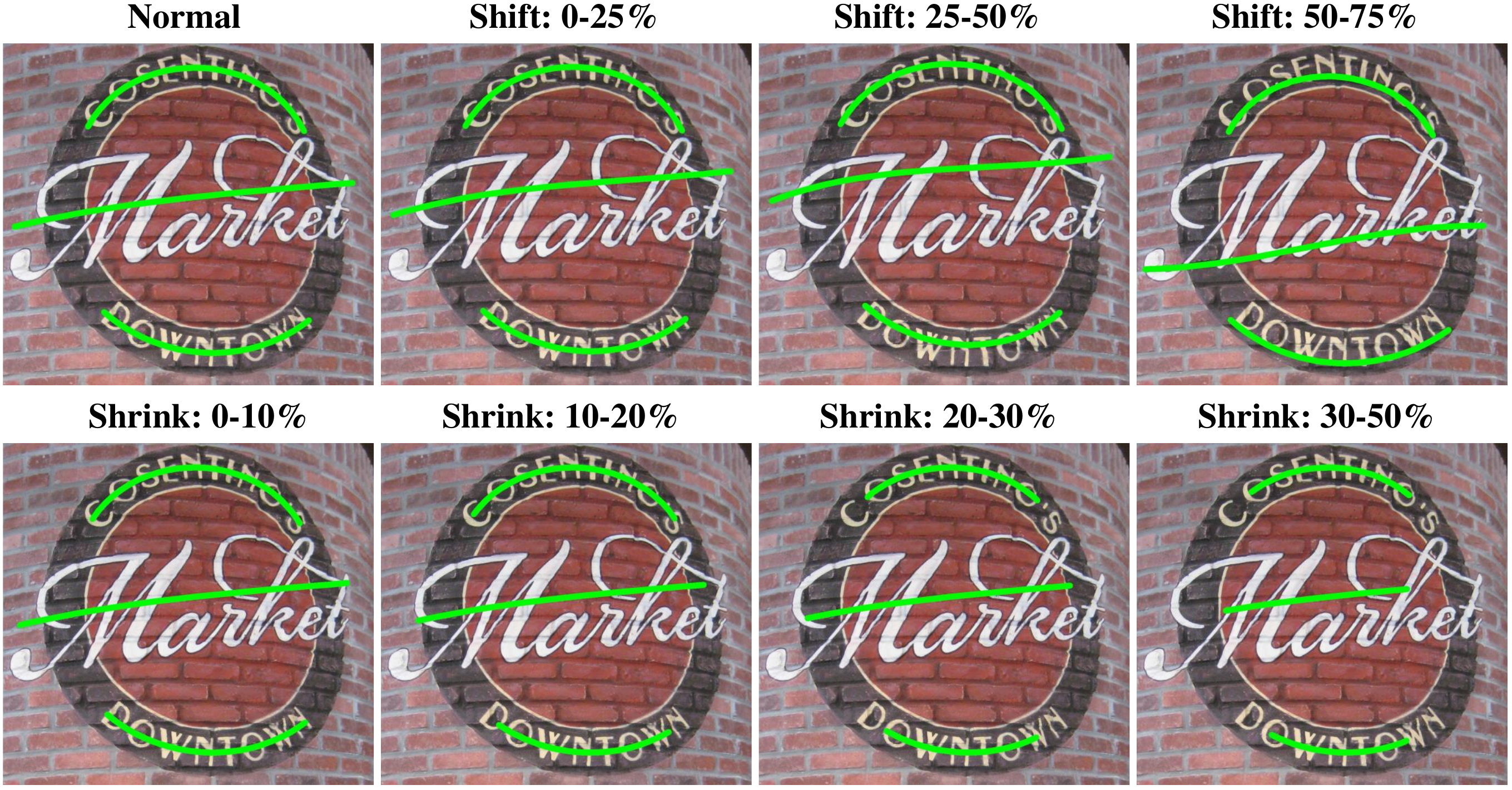}
    \caption{The illustration of line labels at different noisy levels.}
    \label{fig:noisy_label_vis}
\end{figure}

\begin{figure}[!t]
    \centering
    \includegraphics[width=\linewidth]{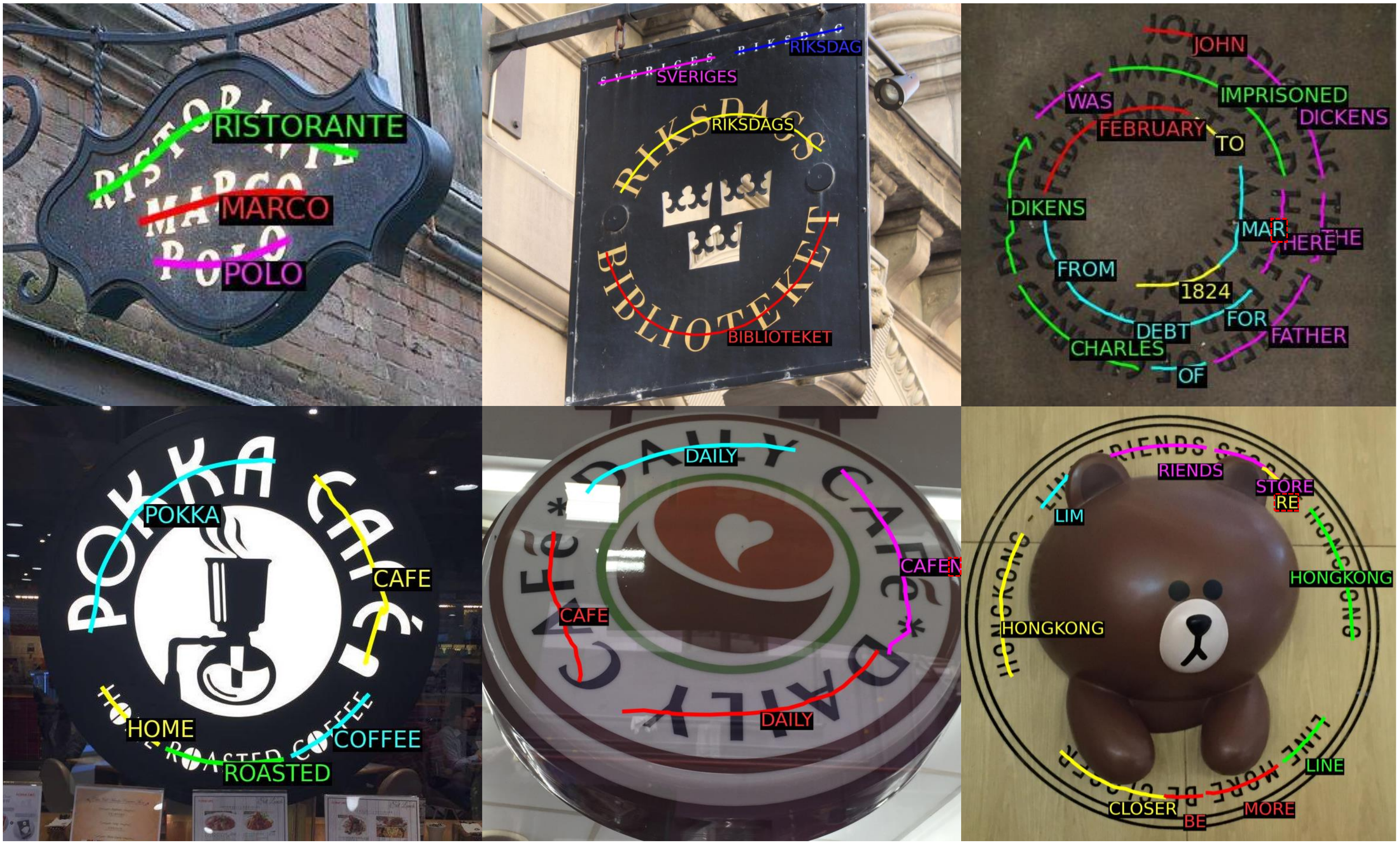}
    \caption{Spotting visualizations of DeepSolo using line annotations. The predicted center curve points in each text instance are connected, forming word lines.}
    \label{fig:line_spotting_vis}
\end{figure}

\section{More Visualizations}
\label{sec:vis}

\begin{figure*}[!t]
    \centering
    \includegraphics[width=\linewidth]{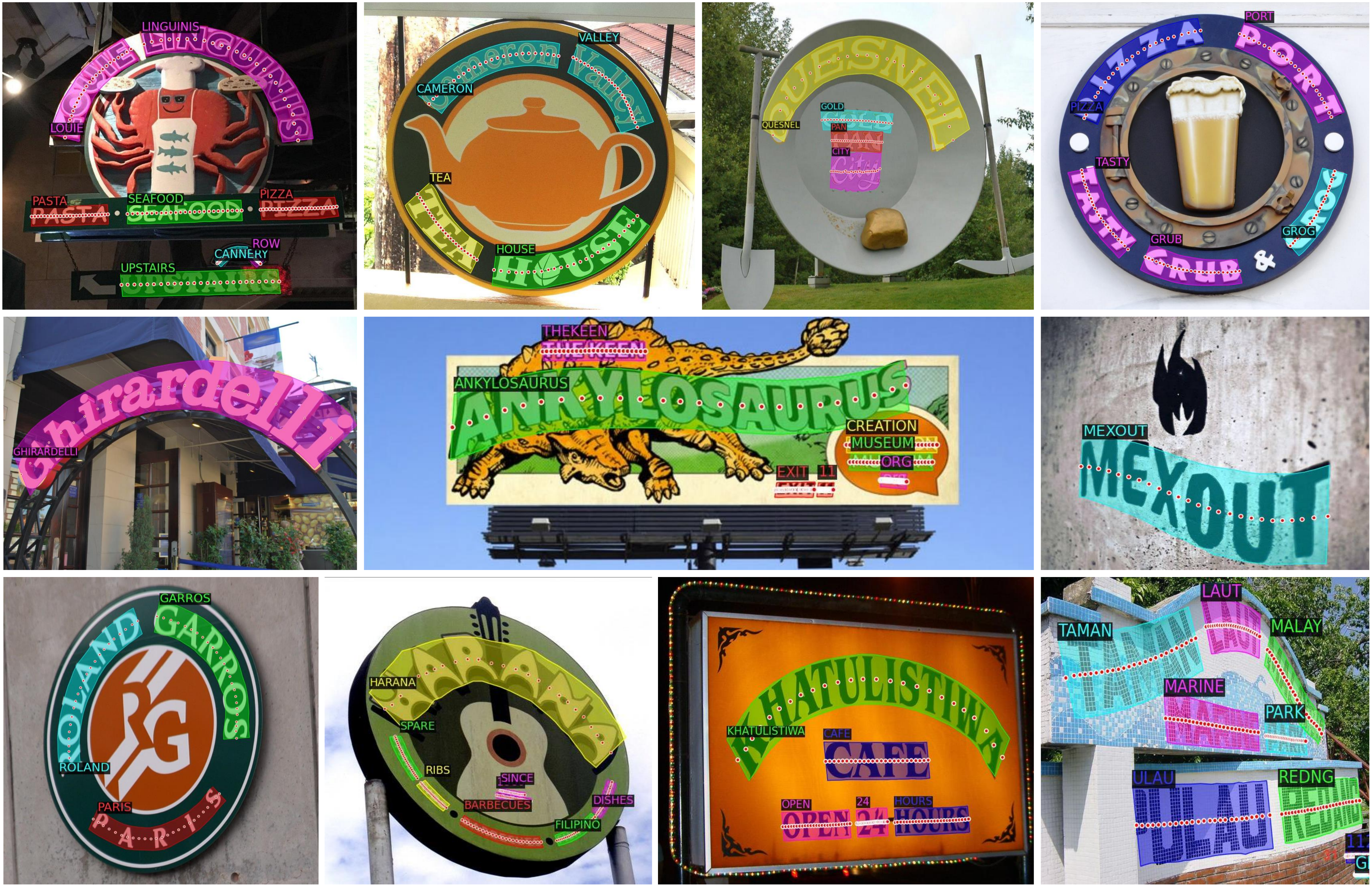}
    \caption{More qualitative results on Total-Text.}
    \label{fig:tt_vis}
\end{figure*}

\begin{figure*}[!t]
    \centering
    \includegraphics[width=\linewidth]{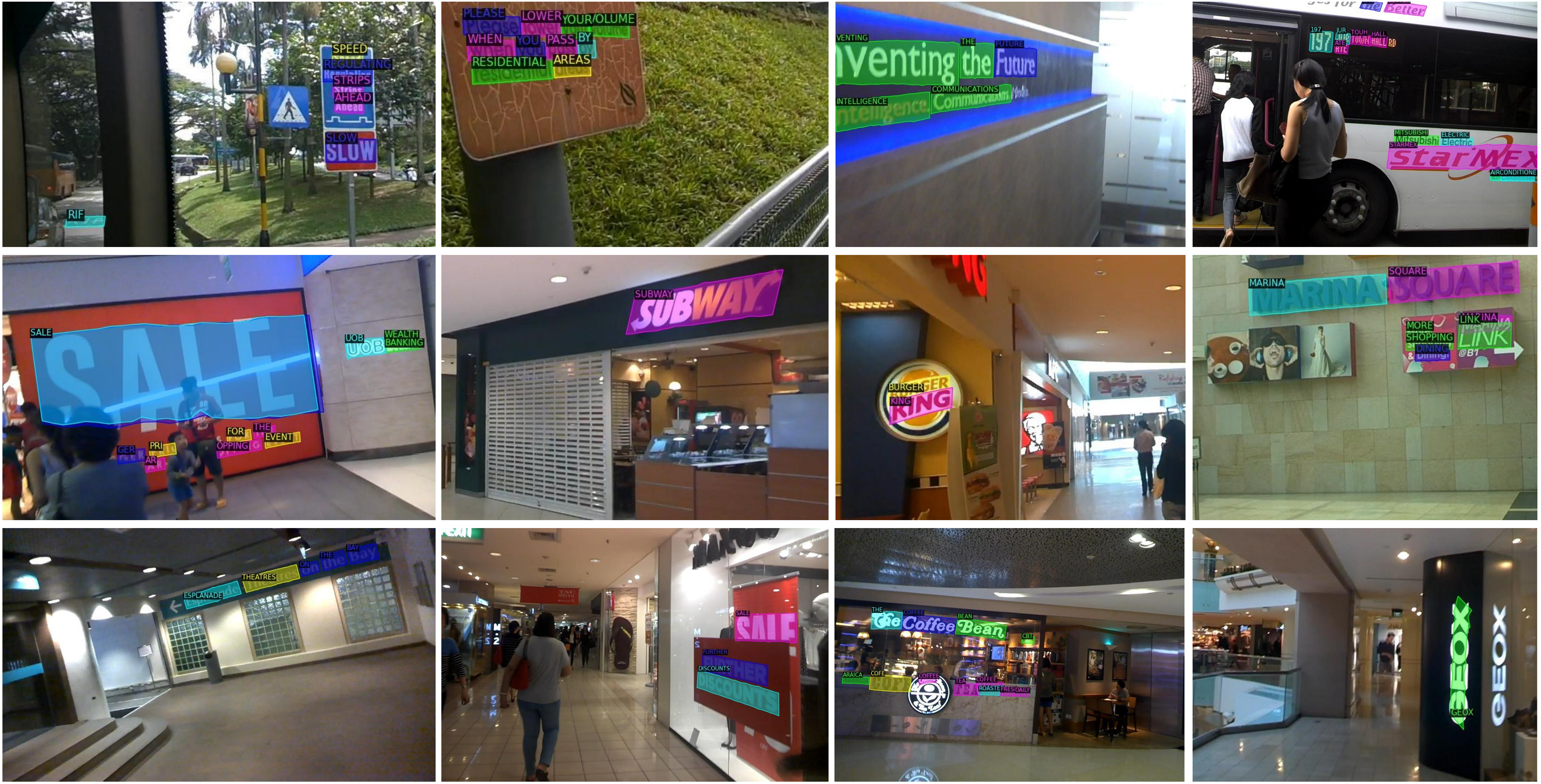}
    \caption{More qualitative results on ICDAR 2015. Center curve points are hidden for better view of small text instances.}
    \label{fig:ic15_vis}
\end{figure*}

\begin{figure*}[!t]
    \centering
    \includegraphics[width=\linewidth]{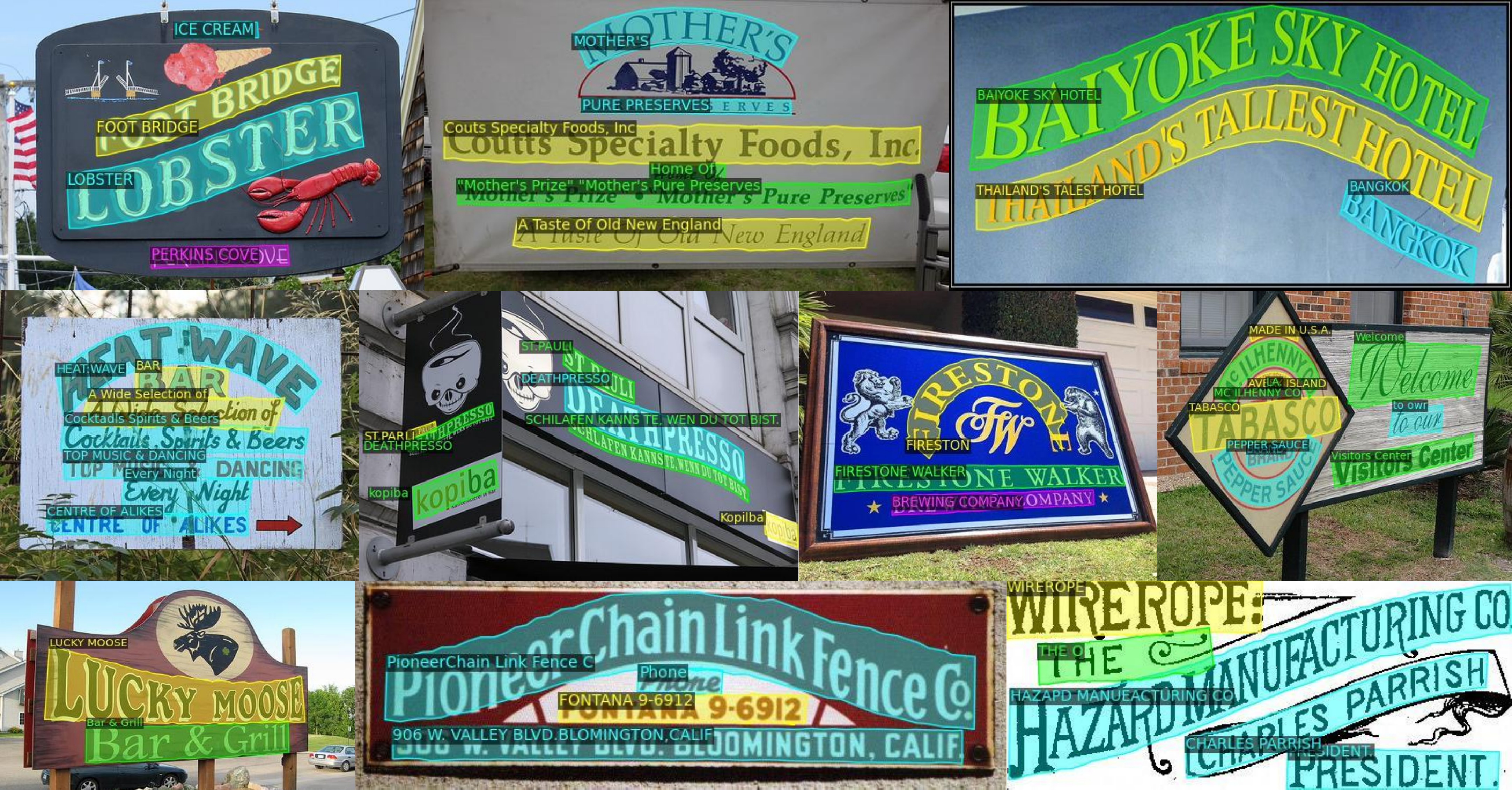}
    \caption{More qualitative results on CTW1500.}
    \label{fig:ctw_vis}
\end{figure*}

\subsection{Visualizations Using Line Annotations}
In Sec. 4.5 of the main paper, we study the model sensitivity to different line locations. Here, we provide a group of visualizations in \cref{fig:noisy_label_vis} to intuitively show the noisy line locations. Moreover, some qualitative spotting results of DeepSolo using normal line annotations are presented in \cref{fig:line_spotting_vis}. Without special design, DeepSolo can correctly recognize most up-side-down text instances with stronger rotation augmentation.

\subsection{More Qualitative Results on Benchmarks}
More qualitative results on Total-Text, ICDAR 2015, and CTW1500 are provided in \cref{fig:tt_vis}, \cref{fig:ic15_vis}, and \cref{fig:ctw_vis}.

\section{Limitation and Discussion}
\label{sec:limitation}
We adopt the label form which is in line with the reading order to implicitly guide DeepSolo to learn the text order. However, when the label form is not in line with the reading order or the predicted order is incorrect, how to get the accurate recognition result is worth further exploration. In this work, we do not utilize an explicit language-aware module to progressively refine recognition results. The combination of DETR-based DeepSolo and language modeling may be promising. Besides, we only study the English scene text spotting framework, we plan to develop a simple and unified multi-language scene text spotter based on DeepSolo.

{\small
\bibliographystyle{ieee_fullname}
\bibliography{egbib}
}

\end{document}